# Application of a semantic segmentation convolutional neural network for accurate automatic detection and mapping of solar photovoltaic arrays in aerial imagery

Joseph Camilo[1], Rui Wang[1], Leslie M. Collins[1], *Senior Member, IEEE,* Kyle Bradbury[2], *Member, IEEE,* and Jordan M. Malof[1], *Member, IEEE*

[1]Department of Electrical & Computer Engineering, Duke University, Durham, NC 27708
[2]Energy Initiative, Duke University, Durham, NC 27708

*Abstract* — We consider the problem of automatically detecting small-scale solar photovoltaic arrays for behind-the-meter energy resource assessment in high resolution aerial imagery. Such algorithms offer a faster and more cost-effective solution to collecting information on distributed solar photovoltaic (PV) arrays, such as their location, capacity, and generated energy. The surface area of PV arrays, a characteristic which can be estimated from aerial imagery, provides an important proxy for array capacity and energy generation. In this work, we employ a state-of-the-art convolutional neural network architecture, called SegNet (Badrinarayanan et. al., 2015), to semantically segment (or map) PV arrays in aerial imagery. This builds on previous work focused on identifying the locations of PV arrays, as opposed to their specific shapes and sizes. We measure the ability of our SegNet implementation to estimate the surface area of PV arrays on a large, publicly available, dataset that has been employed in several previous studies. The results indicate that the SegNet model yields substantial performance improvements with respect to estimating shape and size as compared to a recently proposed convolutional neural network PV detection algorithm.

*Keywords*— solar energy, object detection, image recognition, satellite imagery, photovoltaic

## I. INTRODUCTION

In this work, we consider the problem of developing algorithms that can automatically identify small-scale (or distributed) solar photovoltaic arrays in very high resolution (VHR) aerial imagery, e.g., $\leq$ 0.3m per pixel. This VHR imagery makes it possible to visually identify PV arrays in the imagery and estimate their shapes and sizes. Fig. 1 (left) shows an example of the VHR aerial imagery used in this work, and some manually annotated PV arrays. Once the PV arrays in the imagery are annotated, information such as capacity and energy generation can be estimated at much higher geospatial resolutions than are currently available (e.g., city-level instead of state-level) [1], [2]. In contrast to manual human annotation, this automated approach is faster, cheaper, and generally more scalable.

### A. Previous work and its limitations

The idea of using computer algorithms to automatically detect solar arrays in VHR imagery was first investigated in [1] (on a small-scale dataset) and [2] (on a large scale dataset). These initial PV detection algorithms were designed using traditional image recognition approaches, consisting of hand-crafted image features and supervised classifiers [1], [2]. Recently, convolutional neural networks (CNNs) have yielded substantial improvements over more traditional image recognition approaches on a variety of remote sensing problems [3,4], including PV detection [5,6].

While the CNN-based PV detector proposed in [5] provided excellent detection capabilities, it is limited in its ability to accurately resolve the shape and size of PV arrays. We use the term *mapping* to denote tasks in which the primary objective is precise pixel-wise detection of objects in geospatial data, such as imagery. Fig. 1 presents results from the detector in [5], illustrating its limited mapping capability. This limitation is not a problem for some applications in which the primary goal is simply to detect the presence of individual PV arrays (e.g., counting rooftop installations). However, it does present a problem for other applications, such as array capacity estimation, where accurate shape and size estimation is critical. For example, our group recently demonstrated that the capacity of an array can be accurately estimated if an accurate annotation of VHR imagery of the array (i.e., a mapping) is available [7]. Therefore, there is a strong motivation to develop PV array detectors that can accurately map PV arrays.

### B. Semantic segmentation for PV array mapping

In this work, we propose a new PV array detector that provides accurate PV array mappings. The previous CNN-based detector, proposed in [5], is based on a "VGG" CNN architecture. The VGG architecture was originally designed for image classification tasks, in which the goal is to estimate the probability that a PV array exists somewhere in the input image. This design objective makes the VGG-based network less suitable for precise pixel-wise object recognition in imagery. In this work, we replace the VGG architecture with a recently proposed architecture, "SegNet" [8], that is specifically designed for this task. In the computer vision community, this pixel-wise recognition task is referred to as *semantic segmentation*, or simply segmentation. Fig. 2 presents an illustration of the difference between semantic segmentation (SegNet) and the classification task (VGG).

Note that mapping is a special case in which segmentation is applied to geo-spatial data.

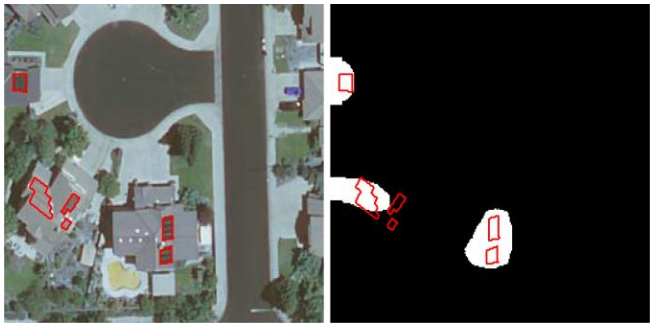

Fig. 1. (left) An example of an orthographic image with the solar PV locations annotated in red, and (right) the predictions made by the previously proposed CNN-based PV detector (in white).

In this work we compare two PV array detectors: one based upon a VGG architecture [5], and new PV array detector employing a SegNet CNN architecture. In order to compare these two detectors, we estimate their pixel-wise (mapping) and object-wise recognition performance on a large collection of publicly available VHR imagery [9] in which all the PV arrays have been annotated. This dataset has been used in several previous publications for the validation of PV array detectors [2], [5]. The results of our experiments indicate that the proposed detector offers substantial improved PV mapping capabilities.

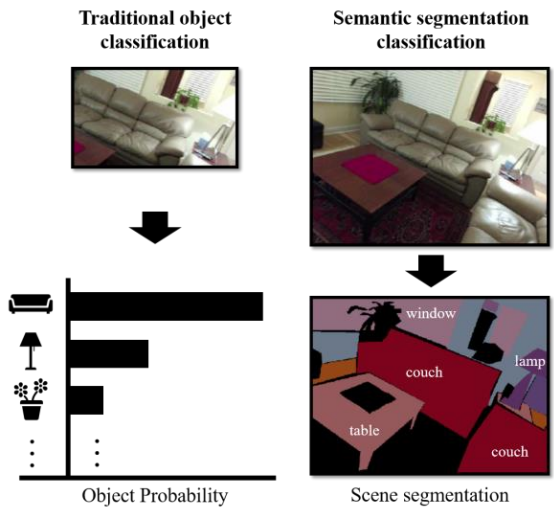

Fig. 2. Illustration of the semantic segmentation problem (right) compared to traditional classification (left). In traditional object classification, the goal is to identify which objects are in the scene. This is contrasted with the semantic segmentation goal of not only identifying which objects are present but also where they are located in the input image. Images from [8].

The concept of using semantic segmentation for extracting the locations of solar PV arrays was first introduced in [6] where the authors present results for a testing area of 6.75 km². We build on that work by enlarging the study area, making the experimental data publicly available, and including a detailed analysis of performance comparison for both pixel-wise and object-wise scoring metrics in the context of previous studies, and using a more modern architecture.

The remainder of this paper is organized as follows. Section II describes the aerial imagery dataset. Section III describes the design and training of the two CNN detectors. Section IV describes the experimental design and results. Lastly, section V describes the conclusions and future work.

## II. AERIAL IMAGERY DATASET

All experiments were conducted on a dataset of color (RGB) aerial imagery, collected over the US city of Fresno, California. All of the imagery is ortho-rectified aerial photography collected in the same month in 2013, with a spatial resolution of 0.3 meters per pixel. The locations of PV arrays in the aerial imagery used as ground truth were manually annotated by human annotators. The dataset used in our experiments (including annotations) is a subset of a larger set of publicly available imagery [9]. The subset of imagery used in our experiments encompasses 135 km² of surface area, and 2,794 PV array annotations.

To avoid a positive bias in the performance evaluation of our algorithms, we split our experimental dataset into two disjoint sets: Fresno Training and Fresno Testing. The data used in these splits is the same as the data used for PV algorithm evaluation in several previous studies [2], [5]. A summary of the imagery in each split is presented below in Table 1, and the unique identifier of each image can be found in [2].

TABLE 1 SUMMARY OF FRESNO COLOR AERIAL IMAGERY DATASET

| Designation | Area of Imagery | Number of images | Number of PV Annotations |
|---|---|---|---|
| Fresno Training | 90 $km^2$ | 40 | 1,780 |
| Fresno Testing | 45 $km^2$ | 20 | 1,014 |

## III. CONVOLUTIONAL NEURAL NETWORKS FOR PV ARRAY DETECTION AND MAPPING

In this section, we provide the details of the two CNN PV array detectors considered in this work: the VGG-based detector from previous studies, and the SegNet-based detector. Both classifiers are designed to take a 41x41 pixel (or roughly 12x12 meters) aerial image, or "patch", as input. As discussed in Section 1, the VGG model was designed for image classification tasks, and therefore it returns a single value, indicating the probability that the input patch contains a PV array somewhere. In contrast, the SegNet model returns a 41x41 probability map indicating the probability that each pixel in the original image corresponds to a PV array.

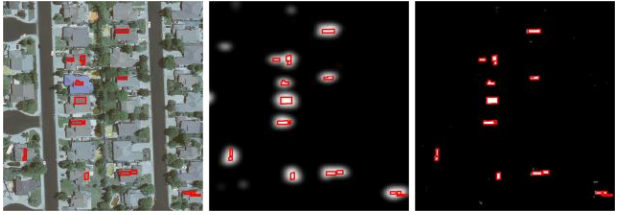

Fig. 3. This is an example of the confidence maps from both detectors, VGG CNN in the middle and SegNet on the right.

Sections A and B present specific design details for the VGG and SegNet CNN architectures, respectively. Specifically, these two sections explain the architectural details and the training procedures for each classifier. In Section C we describe how we aggregate many spatially proximate classifier predictions in order to create large geo-spatial probability maps, such as the ones shown in Fig. 3.

*A. The VGG architecture*

CNNs consist of several processing blocks, or modules, that are applied sequentially to the data. The SegNet and the VGG CNN architectures chosen for this work are illustrated in Fig. 4. The VGG architecture is inspired by the designs of the Visual Geometry Group (VGG) at Oxford [10]. This design consists of two types of modules: "VGG(x)" modules, and fully connected neuron "FC(y)" modules, which are each described below, along with their input parameters, $x$ and $y$.

A VGG module contains two consecutive convolutional filter layers, each with $x$ filters that are 3x3 pixels in size. Each of these two convolutional layers is followed by a rectified linear unit (ReLU) activation. The last part of our VGG layer is a 3x3 pixel max-pooling layer, with a stride (i.e., spatial sampling rate) of 2 pixels.

Fully Connected (FC) modules refer to a single layer of fully connected neurons. These layers have $y$ neurons, and each neuron in a layer is connected to every output from its preceding layer. There are two FC layers in the CNN, which collectively act as a classifier of the output from the preceding convolutional layers. The last layer is a two-way soft-max unit, which returns the probability that the input image patch contains a PV array. An overview of the architecture is illustrated in Fig. 4A, and the specific filter sizes are given in Fig. 5.

Stochastic gradient descent was used for CNN training, similar to [10], [11]. We used a batch size of 100, a learning rate of 0.001, a momentum of 0.9, and a weight decay of 0.0001. Training consisted of 25 epochs (complete passes through the data), and the learning rate was dropped by half the magnitude every 5 epochs. The patches from four of the large training images (10% of the total training imagery) were removed from training and used as a validation dataset.

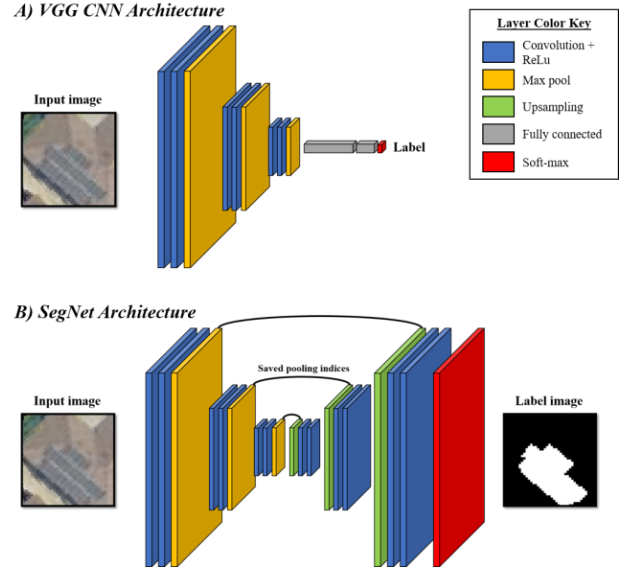

Fig. 4. These are the architectures for the two CNNs used in this work. (A) is the VGG CNN, which has been previously applied to the PV array detection problem. This is a traditional CNN where a single output label is returned for the entire 41x41 image. (B) is the proposed SegNet model, which consists of a series of "upsampling" and "deconvolution" layers that result in a confidence map output.

*B. The proposed SegNet architecture*

The architecture for the second CNN utilized in this work is directly inspired by the SegNet model [8], and is illustrated in Fig. 4B. The first half of this architecture is identical to the *convolutional* layers of the VGG classifier: it has the exact same first three layers. However, instead of two fully connected layers afterwards, there are a series of additional "deconvolutional" layers and "upsampling" layers. These additional layers occur in triplets: one upsampling layer, followed by two consecutive convolutional layers. These triplets are denoted "D-VGG(x)" layers, where "D" stands for deconvolution and x once again denotes the number of convolutional filters in each layer. Each of these D-VGG(x) layers returns a larger feature map than it receives, with the ultimate goal of providing a PV array probability estimate at each pixel location in the original input image.

| VGG CNN | SegNet |
|---|---|
| Input: 41 x 41 RGB image | Input: 41 x 41 RGB image |
| VGG(64) | VGG(64) |
| VGG(128) | VGG(128) |
| VGG(128) | VGG(128) |
| FC(128) | D-VGG(128) |
| FC(32) | D-VGG(128) |
| | D-VGG(64) |
| Output: 2-way soft-max | Output: 2-way soft-max |

Fig. 5. Specific values for the layer definitions in both networks used in this work.

We provide a brief description of the upsampling layer of the SegNet architecture here, but we refer the reader to [8] for further details. Each D-VGG(x) layer is paired with a corresponding VGG(x) layer (see Fig. 4B), and the "upsampling" layers use the max pooling index from its VGG pair to embed a feature value in an expanded feature map. This expansion creates a larger, though sparse, feature map using the features from the preceding feature layer. The last layer of the SegNet network consists of a two-way soft-max layer, which is applied densely to the feature vector at each location in the 41x41 feature maps at the output of the SegNet architecture. The precise details of our SegNet architecture are given in Fig. 5.

Similar to the VGG CNN stochastic gradient descent was used to train the network. We used a batch size of 100, a learning rate of 0.001, a momentum of 0.9, and a weight decay of 0.0005 in accordance with the original SegNet literature [8]. Training consisted of 100 epochs (complete passes through the data), and the learning rate was held constant throughout, again as suggested by the SegNet authors. The same held-out set of training images was used to validate the network as it was training.

*C. Creating large contiguous confidence maps*

Both SegNet and VGG operate on 41x41 input patches, and further, they return output predictions of varying size: a single output from VGG and a 41x41 output from SegNet. Ultimately, we need dense predictions over large contiguous spatial regions. To obtain such maps, we apply the networks to 41x41 patches that are extracted densely from the larger set of aerial imagery, and then "stitch" their outputs together to create larger "probability maps". Fig. 6 presents the details of this stitching process. The resulting probability maps can be treated as a list of detected objects (one for each pixel) and the corresponding probability or confidence, $c$, that the object is a panel. This information can be combined with ground truth label maps that indicate which pixels are truly located over panels, in order to score the pixel-wise output of the detector.

*D. Detecting individual PV array objects*

The prediction maps described in Section 3.3 provide pixel-wise estimates of panels, however, in many practical applications it is desirable to identify individual PV arrays. We achieve this by labeling any pixel with a confidence, $c \geq 0.5$, as a panel pixel, and all other pixels are labeled as non-panel. Recall that a $c$ simply represents the predicted probability of that pixel to correspond to a panel. Therefore, $c \geq 0.5$ indicates that the model considers the pixel more likely to be a panel than a non-panel. This approach is similar to labeling strategies in many other semantic segmentation applications [8].

This thresholding operation results in a binary image, in which pixels are labeled as either a panel (label value of one), or non-panels (label value of zero). In order to identify individual arrays in this binary image, we identify connected components (i.e., contiguous groups) of panel pixels. Any such grouping of panel pixels is returned by the detector as a PV array detection. In order to measure the accuracy of these predictions we need a criterion for deciding when an object returned by the detector is close enough, and similar enough (e.g., in shape and size), to a true PV array annotation in order to consider it a correct detection (versus a false detection). The criterion for labeling detections as correct or incorrect is discussed in Section IV.C. In order to evaluate the PV detections, we will also need to assign a probability estimate, indicating the relative likelihood that the detected object is a PV array. This object probability is obtained by averaging the probabilities of the pixels within the object. The result of this processing is (i) a list of detected PV array objects, (ii) their associated probability values, and (iii) labels indicating whether the detection is actually a solar array or not. This is analogous to the pixel-wise detector outputs, and can be used in a similar manner to score the object-wise performance of the detector.

## IV. EXPERIMENTAL DESIGN

In this section, we describe the experimental design used to evaluate the classification performance of the two PV array detectors described in Section III. Broadly, the design consists of training the detectors using the 'Fresno Training' imagery, and then evaluating the performance of the trained detectors on the 'Fresno Testing' imagery.

*A. Training data extraction and augmentation*

To train the two CNN models, we need to extract a large number image patches corresponding to each class (i.e., PV array versus non-PV). There are over a billion pixels in the Fresno Training imagery, and one could conceivably extract a training patch at every location. However, due to computational constraints, we extract roughly 2 million total training patches. We use the same large images and procedure for extracting training images as the previous VGG CNN study [5].

To obtain non-PV training images, we first sampled every fifth pixel in the 'Fresno Training' aerial imagery, retaining only those locations that did not intersect with a PV annotation. Of the remaining pixel locations, roughly 1.5 million (75% of 2 million) patches were randomly sampled to be included in CNN training. To obtain training images containing PV arrays, we sampled every 3rd pixel of the 'Fresno Training' aerial imagery, and retained only pixel locations that intersected with PV array annotations. To reduce the redundancy of these densely sample patches, we randomly sampled only 30% of the total available solar array patches. Each of these retained patches was then copied four times, and a random rotation was applied to each of the four copies to further augment the training data. This yielded roughly 500,000 panel training patches (25% of 2 million).

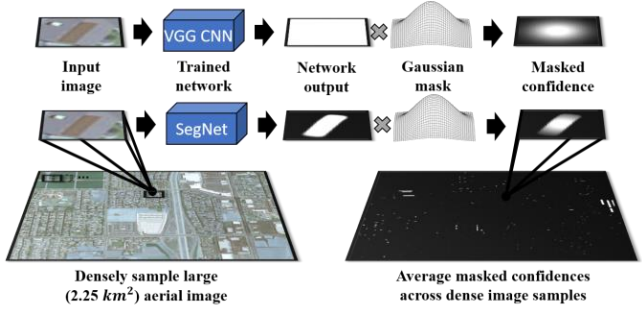

Fig. 6. Illustration of the process of creating large contiguous prediction maps using the (a) VGG and (b) SegNet detectors. In both cases, patches (41x41) are extracted densely, with a stride of 10, from the larger aerial imagery. A prediction is obtained for each patch, and the VGG output is upsampled to the same size as the SegNet output. These prediction maps are then placed into the original location of the input patch, which results in multiple predictions at each spatial location. These predictions are combined through a weighted average, wherein the relative weighting of each prediction is obtained from a Gaussian window applied to each 41x41 output image. This tends to give lower weight to predictions that were made towards the edge of their respective patch. We found that this weighted average yielded higher quality predictions. Note that in the case of the VGG network, this weighting corresponds to interpolation with a Gaussian kernel [18].

*B. Pixel-based versus object-based scoring*

Two types of scoring are utilized to compare the performance of the PV array detectors. The first is *pixel-based* scoring which treats every pixel in the aerial images as a detection returned by the detector. Both methods for scoring classification performance rely on an algorithm's ability to assign greater panel probabilities to true panel locations, while providing low probabilities to the pixel location which do not correspond to true panel locations. Recall that ground truth PV pixel locations are provided via manual labeling of the aerial imagery.

While pixel-based scoring is a popular criterion for evaluating many semantic segmentation algorithms [8], since their design objective is to obtain accurate pixel-wise labeling of imagery, some remote sensing applications may not require highly accurate pixel-wise labeling if individual objects (e.g., panels) are reliably identified. We can consider the application of counting individual rooftop PV array installations, which can be achieved with relatively poor pixel-wise performance. An example of such an outcome is illustrated for PV array detection in Fig. 7.

The main advantage of object-based scoring over pixel-wise scoring is that it ensures that the detector outputs are spatially contiguous, and can therefore be attributed to one real-world object. In contrast, a detector might achieve a high pixel-wise performance, but if the detections consist of many small connected regions (perhaps over a single real-world object) this can result in very poor object-based performance. Indeed, such a detector would be very inaccurate for counting PV arrays, because it would obtain many small detections over one array, returning a positively biased estimate of the number of individual PV array installations.

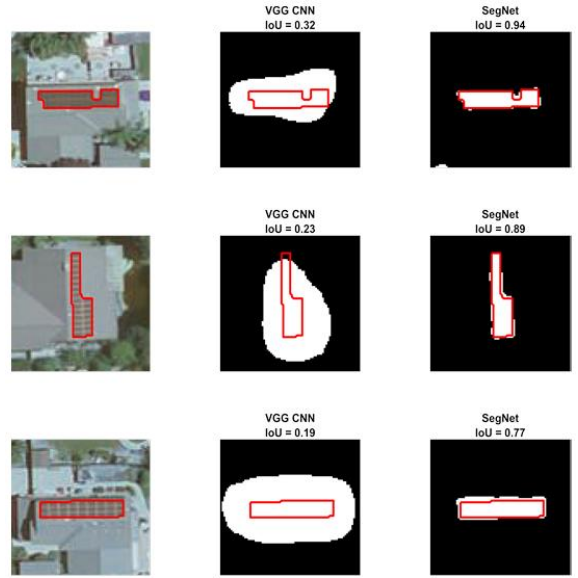

Fig. 7. Examples of the detection regions used in object-based scoring, with the computed IoU value with the corresponding truth boundary. The SegNet predicts the shape of the PV array consistently better than the VGG CNN.

It is also possible to combine the criterion used in pixel-based and object-based into one more stringent scoring criterion. This can be achieved by increasing the degree to which a detected objected must match with a true PV array annotation in order to be considered a correct detection. In the limit, once could require that the detected object match perfectly with one true PV annotation. This would require perfect pixel-wise detection and connectedness among the detected objects. In the next section, we discuss a common metric for evaluating the shape and size similarity between two objects, and how we employ it for object-based scoring.

*C. Criteria for considering a detected PV array a correct detection*

In order to measure the shape and size similarity between two image objects (i.e., contiguous groups of pixels) we use the *intersection over union* (IoU) Index. More specifically, we use IoU to measure the similarity between detections and ground truth annotations, in order to determine when a detected region will be considered a correct detection. The IoU measure is common for scoring semantic segmentation algorithms in remote sensing applications [8], [12]. Given two sets of pixels (i.e., regions), denoted by A and B, IoU is given by

$$\text{IoU} = \frac{|A \cap B|}{|A \cup B|}. \quad (1)$$

The IoU value generally increases as two regions become more similar. This works utilizes multiple thresholds of IoU when considering the object-based scoring for PV array detection. By altering the IoU threshold, a detector's ability to resolve precise panel shape and size can be evaluated. In Fig. 7 there are three examples of the detection regions from both the VGG CNN and the SegNet models. Each has the computed IoU value between the detection region and the truth boundary (in red). Notice on these preliminary results that the SegNet model has much higher IoU values, which implies that the SegNet model is better able to predict the shape and size of the PV arrays.

*D. Scoring metrics*

Whether pixel-based or object-based scoring is employed, the output of the detectors is comprised of (i) a list of detections, (ii) each with an assigned a probability value, and (iii) its true class (i.e., PV array or non-PV). With this information, it is possible to construct a precision-recall (PR) curve. The PR curve is an established performance metric for object detection in aerial imagery [13]–[16], and therefore it is adopted here. PR curves measure the performance tradeoff between making correct detections and false detections, as the sensitivity of a detector is varied. The sensitivity of the detector is controlled by a threshold parameter, $\tau$, which is applied to determine whether a pixel or object will actually be returned as a detection by the detector. All detections with a probability greater than $\tau$ are returned, while all others are discarded.

In order to compare multiple PR curves across a variety of experiments, a single metric is computed to summarize the PR curve. The summary metric used here is known as the maximum $F_1$ score [17]. An $F_1$ score is a measure of performance at any point along the PR curve, and is defined as the harmonic mean of the precision and recall for a given detector operating point. The maximum $F_1$ score attempts to summarize a given PR curve by returning the maximum $F_1$ score over all possible operating points. The maximum $F_1$ is given by (2).

$$Max\ F_1 = \max_{\tau} 2 \frac{P(\tau)R(\tau)}{P(\tau) + R(\tau)}. \qquad (2)$$

Here $P(\tau)$ and $R(\tau)$ are the precision and recall, at a given detector threshold, $\tau$. The $F_1$ score ranges from 0 to 1, where 1 corresponds to perfect precision at a recall of 100%. With the $F_1$ scores computed across the range of detector thresholds the max value is selected as a single metric to summarize the PR curve.

*E. Pixel-based performance results*

In this section, the pixel-based results are presented for both detectors. In Fig. 8 are the PR curves for pixel-based scoring on the testing images for both the VGG CNN and the SegNet models. The SegNet model outperforms the VGG CNN in pixel-based scoring across the whole PR curve. These results indicate that SegNet is substantially more effective for pixel-wise accuracy. As we will see in the next section, this performance advantage extends to *most* object-based scoring criteria as well.

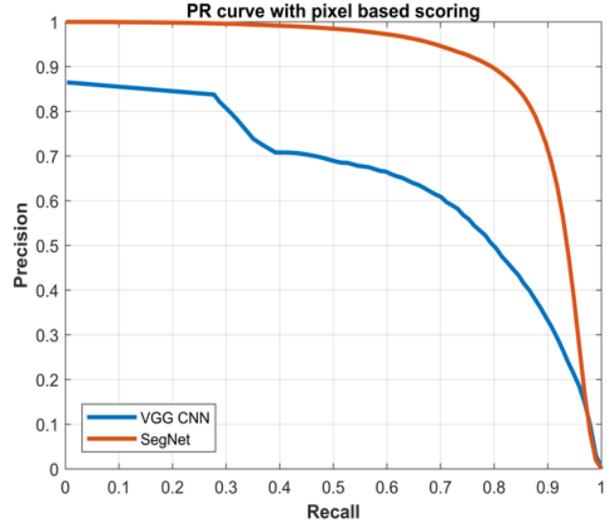

Fig. 8. PV array detection performance results using the pixel-based scoring method for both the SegNet and VGG CCN models.

*F. Object-based performance results*

This section includes the description of object-based scoring and the detection performance results from the two CNN models. For this purpose, we again use precision-recall (PR) curves. Found in Fig. 9 are the PR curves for the two networks with two different IoU thresholds, 0.5 and 0.1. Recall from the previous section that a IoU threshold of 0.1 requires relatively little shape and size similarity between the detected object and the ground truth annotation. This is primary reason for the performance drop off of the VGG CNN when increasing the IoU threshold for object based detection.

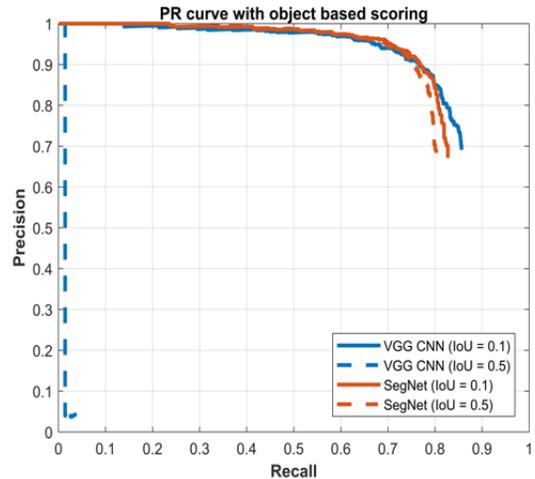

Fig. 9. Results of running two different network architectures, VGG CNN and SegNet, on the 'Fresno Testing' dataset. This is the object based performance for two different IoU thresholds, 0.5 and 0.1. While there is similar Pv array detection performance for when the IoU threshold is 0.1 but when only considering detection with a IoU greater than 0.5 the VGG CNN performance falls dramatically.

Additional IoU thresholds were tested in order to comprehensively evaluate the ability of SegNet to resolve individual PV array objects, as well as their precise shape and size. The max $F_1$ scores for a range of IoU thresholds are own in Fig. 10. The SegNet model has better detection performance as the IoU threshold approaches 0.6 and then begins to drop rapidly. The VGG CNN drops in performance immediately when increasing the IoU threshold to be above 0.1. This once again suggests that the SegNet is predicting PV array shape with much greater accuracy than the previous VGG CNN model. These results also provide a measure of SegNet's PV mapping capabilities for future comparison.

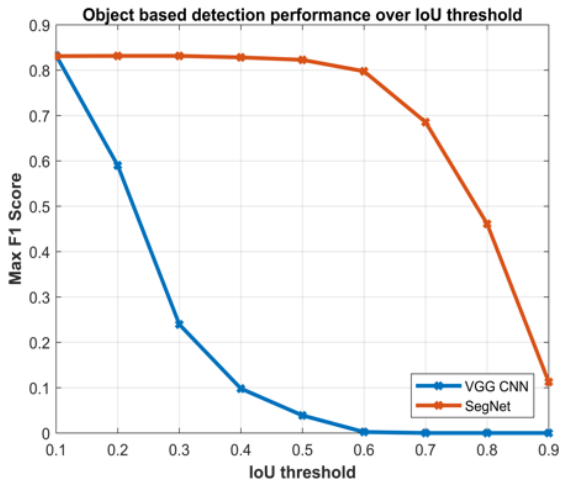

Fig. 10. PV array object-based detection results from both the VGG CNN and the SegNet models as the IoU threshold is varied. Here performance is measured by the max $F_1$ score across the individual PR curves.

## V. CONCLUSIONS

In this work, we investigated the use of a semantic segmentation CNN for the problem of detecting solar PV arrays in aerial imagery. This network, called SegNet, was evaluated and compared with a previous CNN using a large dataset of publicly available aerial imagery, encompassing roughly $135 km^2$ of surface area. Two performance metrics were used to compare detectors, pixel-based and object-based scoring. In pixel-based scoring, the SegNet model substantially outperformed the previously published CNN. In object-based scoring the measure of similarity between detection regions and ground truth annotations was varied to quantify each networks' ability to detect shape and localization of PV arrays. Results from the object-based scoring strongly suggest that the SegNet model is substantially better at predicting the shape and location of the PV arrays, which is a significant step towards producing accurate estimates of power capacity and energy generated by solar PV arrays directly from satellite imagery.

ACKNOWLEDGEMENTS

We would like to thank the NVIDIA Corporation, for donating the graphics processing unit (GPU) for this work. We would like to thank the Duke University Energy Initiative for their support for this work.